# Long-term face tracking in the wild using deep learning

Kunlei Zhang[a], Elaheh Barati[a], Elaheh Rashedi[a,*], Xue-wen Chen[a]

[a]*Dept. of Computer Science, Wayne State University, Detroit, MI*

**Abstract**

This paper investigates long-term face tracking of a specific person given his/her face image in a single frame as query in a video stream. Through taking advantage of pre-trained deep learning models on big data, a novel system is developed for accurate video face tracking in the unconstrained environments depicting various people and objects moving in and out of the frame. In the proposed system, we present a detection-verification-tracking method (dubbed as 'DVT') which accomplishes the long-term face tracking task through the collaboration of face detection, face verification, and (short-term) face tracking. An offline trained detector based on cascaded convolutional neural networks localizes all faces appeared in the frames, and an offline trained face verifier based on deep convolutional neural networks and similarity metric learning decides if any face or which face corresponds to the query person. An online trained tracker follows the face from frame to frame. When validated on a sitcom episode and a TV show, the DVT method outperforms tracking-learning-detection (TLD) and face-TLD in terms of recall and precision. The proposed system is also tested on many other types of videos and shows very promising results.

## 1. Introduction

Consider a video stream taken in unconstrained environments depicting various people and objects moving in and out of the camera's field of view. Given a bounding box defining a face of a specific person in a single frame, the goal is to automatically determine his/her face's bounding box or indicate that this person is available or not in the rest frames of the video. The desired output is the person's faces and the corresponding time slot when he/she appears in the video. This task is referred to as long-term face tracking [1, 2, 3].

---

*Rashedi and Barati contributed equally to this article.
**August 2016, San Francisco, CA, USA

Long-term face tracking is an appealing research direction with increasing demands. For example in the era of social networking, when more and more videos are continuously uploaded to the Internet via video blogs, social networking websites, face tracking technology can track and retrieve all the shots containing a particular celebrity from thousands of short videos captured by a digital camera; or it can locate and track suspects from masses of city surveillance videos (e.g., Boston marathon bombings event).

Long-term real-time tracking of human faces in the wild is a challenging problem because the video may include frame cuts, sudden appearance changes, long-lasting occlusions, etc. This requires the tracking system to be robust and invariant to such unconstrained changes. Since most of tracking methods [4, 5, 6] in the literature have been aimed at the videos in which the target person is visible in every frame, these methods cannot easily handle the long-term tracking situations.

In this work, we develop a new system for accurate long-term video face tracking in the wild by taking advantage of pre-trained deep learning models on big data. The main idea is based on a detection-verification-tracking (DVT) method in which we propose to decompose the long-term face tracking task into a sequence of face detection, face verification, and (short-term) face tracking. Specifically, given a query face of a specific person, the offline pre-trained detector based on cascaded convolutional neural networks localizes all faces appeared in the frames, the offline pre-trained face verifier based on deep convolutional neural networks and similarity metric learning decides if any face or which face corresponds to the query person, and the online trained tracker follows the verified face from frame to frame. The system repeats this procedure until the end of a video. To speed up the system or even make it to be real-time, we can skip a number of frames in two cases: when each short-term tracking is done, and when no face in the current frame is verified to belong to the query person. Since we apply deep convolutional neural networks trained on big data in the wild to face detection and face verification, the system is able to tackle videos taken in unconstrained conditions. Figure 1. provides an overall flowchart of the proposed system which will be described in details in Section 3.

Our main contributions in this paper are two-fold. (i) A DVT method is presented which accomplishes the long-term face tracking task through the collaboration of face detection, face verification, and (short-term) face tracking. To the best of our knowledge, the face verification is, for the first time, performed to guide the (long-term) face tracking. (ii) Built on the DVT method, a novel and accurate long-term face tracking system is designed and developed, which can handle various types of video in the wild. Also, this system is an end-to-end one.

The rest of the paper is organized as follows. Section 2 gives a brief introduction on related works on face tracking. The proposed DVT method and the developed system is elaborated in Section 3, which is followed by the experimental results in Section 4. Finally, this paper is concluded in Section 5.

## 2. related work

Many approaches have performed detection to improve the tracking procedure while some of them used offline trained detector [7, 8], and some others used online learned detectors [1, 9, 10, 2]. For example, in [7] the object tracking algorithm applied a detection strategy to validate the tracking results. If the validation was failed, the whole frame would have to be searched again to find the target. Another example of tracking with offline detector [8] employed a detection strategy with particle filtering to improve the tracking algorithm. While these methods utilized pretrained detectors, adaptive discriminative trackers with an online learned detectors to distinguish the target from the background were presented in [9, 10]. Although these methods achieves promising performance in continuous tracking, but if the target leaves the scene slowly and gradually then there will a probability that the procedure may lose tracking the main target and replace it with a wrong subject.

Tracking-learning-detection (TLD) [1] is a method to tackle long-term object tracking in video. In TLD, starting from a single frame, the tracking procedure exchanged information with an online learned detector while the two procedures worked independently. By using a randomized forest as classifier the decision boundary between the object and its background can be represented. In [2], the TLD framework was specified to the application of long-term human face tracking. A validator was employed to decide whether a face patch corresponds to the query face or not. The method used the frontal face detector algorithm [11], and on top of that a module was incorporated to analyze a face patch as a validator. The output was a confidence level which indicated the correspondence of the patch to the specific face. The validator was performed on a collection of example frames which was initialized by a single example in one frame and then extended during tracking by inserting more examples.

A tracking framework presented in [12] combined tracking and detection to support precision and efficiency of tracking under heavy occlusion conditions. Two different strategies based on TLD and wider search window approaches were used for detection. Objects in tracking were represented by sparse representations learned online with update.

Similarly, in this work we also use face detection to improve the tracking procedure. Furthermore, we propose to perform face verification as a validator to guide the tracking. More importantly, we take advantage of deep learning models for face detection and face verification in our system, which enables high accuracy in tracking.

## 3. Methodology

In this section, we will present the proposed DVT method and the developed system for long-term face tracking.

### 3.1. Detection-Verification-Tracking (DVT)

We utilize face detection and face verification to improve the tracking procedure. In the DVT method, we propose to decompose the problem of long-term face tracking into a sequence of face detection, face verification, and (short-term) face tracking. Specifically, given a query face of a specific person, the offline pre-trained detector based on cascaded convolutional neural networks localizes all faces appeared in the frames, the offline pre-trained face verifier based on deep convolutional neural networks and similarity metric learning decides if any face or which face corresponds to the query person, and the online trained tracker follows the verified face from frame to frame.

#### 3.1.1. Face Detection

Recently, the deep learning techniques have revolutionized the performance of face detection. A survey on the most successful face detection methods is given in [13]. Although several state-of-the-art face detection methods reached almost perfect accuracy, they are not fast enough to be suitable for real-time applications. For example, a novel deep learning convolutional network for face detection in [14] achieved high accuracy but was computationally intensive and comparably slow. This impedes its use to real-time purposes like online video tracking.

Considering a balance between effectiveness and efficiency, we need a detection algorithm which not only works well under unconstrained circumstances but also performs at an acceptable speed. Here, we choose a convolutional neural network (ConvNet) detection algorithm, named a cascaded-CNN [15], that can achieve high accuracy with a fast speed. This cascade-CNN includes 6 ConvNets worked in a cascaded way in 3 stages. In each stage, one ConvNet is used for detecting faces vs. non-faces and the other ConvNet is used for bounding box calibration. The output of one stage is used to adjust the detection window position which will be input to the subsequent stage. This method reduces the number of face candidates at later stages by using a ConvNet-based calibration after each detection. More details can be found in [15].

Figure 1: Overall flowchart of the proposed system for long-term face tracking. The input is given as a cropped face which is first detected and then tracked for five frames. Feature query is the average of five feature vectors which are obtained by applying a pre-trained deep convolutional neural network to the five detected faces on the five frames. Tracking continues until there is no frame or the query person disappears from the scene. When he/she appears again, after detection and verification, tracking will be started. The procedure repeats until the end of a video. Ideally, the output is all tracked faces of the query person and their corresponding time slots.

3.1.2. Face Verification

In recent years, CNN based face recognition techniques have obtained a near perfect verification accuracy on some datasets [16, 17, 18]. VGG-face net [16] investigates CNN architecture for face identification and verification with a deep network in the sense that a long sequence of convolutional layers is used. This CNN was trained on 2.6 million face images from more than 2600 people and achieved comparable verification accuracy with the state-of-the-art methods on benchmark data sets. The pre-trained CNN model is also publicly available from this link[1].

In this work, we take advantage of the pre-trained model of VGG-face network to extract features for faces. Specifically, the detected faces are first preprocessed in the same way as in [16], and then we apply the VGG-face CNN to the faces and take the output of the last fully connected layer (without the nonlinearity) as feature representations each of which is a 4096-dimensional vector. Thereafter, we consider the query feature vector and the feature vector of a detect face as a pair of feature vectors. Cosine similarity metric learning is used to verify a pair of features to belong to the same face or different faces.

---

[1] http://www.robots.ox.ac.uk/~vgg/software/vgg_face

In the TLD method [1] the frames are treated as to be independent and the whole frame is being scanned to detect the target. Unlike the TLD method, our verification strategy make the assumption that consecutive frames are related to each other to some extent, therefore the scanning process is performed around the area where the latest detected bounding box was located. This assumption decrease the scanning process time in comparison to TLD.

### 3.1.3. Face Tracking

Most tracking algorithms employ a bounding box given in the first frame and continue tracking based on the initial bounding box. Despite the fact that researchers have been making progress in this field, it still remains highly challenging to design a tracker which can handle all various situations, such as object deformations, illumination changes, fast motions, occlusion, and background clutters, etc. Furthermore, another big challenge of tracking is to handle long-term tracking situations, in which tracking algorithm will continuously confront different conditions as the target may leave the scene and re-appears later.

Most of the tracking algorithms use only one bounding box (or patch) to be tracked [19, 20]. In this work, we employ the reliable patch tracker (RPT) method [4] which identifies and exploits multiple reliable patches instead of only one, where the reliable patches can be tracked effectively through the tracking procedure. With the collaborative use of face detection and verification, the RPT method can handle long-term tracking under the assumption that the object's motion between consecutive frames is limited and whenever the object leaves the scene the verification procedure will stop the tracking.

### 3.2. System Framework

The goal is to design and implement a system which can track a specified human face in an unconstrained video with the long-term setting. Algorithm 1 gives the pseudo code for this system framework. Details on the work flow of the system are described as follows.

The system starts by asking the user to select a face by cropping it in some frame of the video. The cascade-CNN based face detection (see Section 3.1.1) is then performed on the cropped face to obtain a bounding box which indicates the position of the target face in the selected frame. The bounding box is fed into the RPT based tracking algorithm (see Section 3.1.3). At this point, the tracking continues only for five frames to create a sequence of five face images. After the preprocessing (i.e., resize the images to be 224x224x3 and subtract a mean image) of the five face images, the VGG-face ConvNet (see Section 3.1.2) is applied to the five images to produce five feature vectors. The dimension of each feature vector is 4096. We consider the average of the five extracted feature vectors as a feature query (of the query face).

Thereafter, the tracking continues until there is a significant difference in the distance of the position of target face in 2 two-consecutive frames. If the distance difference is significant, the tracking is stopped and face verification on the face in the latest frame is performed. In order to conduct face verification, the feature representation of the face in the latest frame is extracted using VGG-face ConvNet as before and then is compared to the feature query using the cosine similarity metric. If the cosine similarity score is larger than a predefined threshold, tracking continues; otherwise, tracking stops.

The system then moves on to the face detection procedure. Face detection is applied to the whole frame to find all possible human faces. Subsequently, the system performs face verification again on each of the detected faces as aforementioned. If the cosine similarity between one of the detected faces and the query face is higher than the threshold, the tracking continues; otherwise, a number of following frames are skipped and the face detection procedure is conducted again. The skipping number of frame can be defined based on the video type and the frame rate of the video. For example, if faces in the video change fast and move fast, a small value should be set for the skipping number of frame; otherwise, a larger value should be set.

Thus, a sequence of detection, verification and tracking will repeat until the end of a video. All tracked faces of the query person and their corresponding time slots are the output. In our system demo, we show the output in the video with highlighted parts where tracked faces have been appeared in the frames.

**Algorithm 1** Long-term Video Face Tracking.
___
1: Video ← Read sample video
2: similarity-threshold ← Set to a predefined value based on desired level of similarity
3: distance-threshold ← Set to a predefined value
4: skip-frame ← set to a predefined value based on the length of video
5: continue-tracking ← True: A flag that indicates whether to stop or continue tracking
6: if-reappear ← False: A flag that indicates whether the face reappears in proceeding frames or not
7: $f\#$ ← Get the number of the frame where the target face exists: defined by user
8: initial-bounding-box ← Get the position box of the target face in a specific frame from the user
9: detected-face[$f\#$] ← DETECT-FACE(initial-bounding-box, $f\#$)
10: **for** $i = f\# : f\# + 5$ **do**
11:     detected-face[$i+1$] ← TRACK-ONE-FRAME (detected-face[$i$] , $f\#$)
12:     feature-vectors[$i$] ← EXTRACT-FEATURE (detected-face[$i$])
13:     $f\# \leftarrow f\# + 1$
14: **end for**
15: query ← Calculate the average of feature-vectors
16: **while** hasFrame(Video) **do**
17:     **while** continue-tracking == True **do**
18:         detected-face[$f\#$+1]←TRACK-ONE-FRAME(detected-face[$f\#$],$f\#$)
19:         **if** DISTANCE(detected-face[$f\#$],detected-face[$f\#$+1]) > distance-threshold **then**
20:             continue-tracking ← False
21:         **end if**
22:         $f\# \leftarrow f\# + 1$
23:     **end while**
24:     feature-vector ← EXTRACT-FEATURE(detected-face[$f\#$])
25:     cosine-score ← COSINE-SIMILARITY(feature-vector, query)
26:     **if** cosine-score > similarity-threshold **then**
27:         continue-tracking ← True
28:     **else**
29:         face-list ← DETECT-ALL-FACES(Video, $f\#$)
30:         **if** face-list is not empty **then**
31:             **for** face ∈ face-list **do**
32:                 feature-vector ← EXTRACT-FEATURE($face$)
33:                 cosine-score ← COSINE-SIMILARITY(feature-vector, query)
34:                 **if** cosine-score > similarity-threshold **then**
35:                       continue-tracking ← True
36:                     detected-face[$f\#$] ← face
37:                     if-reappear ← True
38:                     break the loop
39:                 **end if**
40:             **end for**
41:             **if** if-reappear == false **then**
42:                 $f\# \leftarrow f\#+$ skip-frame
43:                 $frame \leftarrow$ readFrame(Video, $f\#$)
44:             **end if**
45:         **else**
46:             $f\# \leftarrow f\# + skip - frame$
47:             frame ← readFrame (Video, $f\#$)
48:         **end if**
49:     **end if**
50: **end while**

## 4. experiments and results

This section presents the implementation of the system, the experiments and the evaluation of tracking performances. The proposed DVT method was implemented in Matlab using single thread without further optimization. The graphical user interface (GUI) of the system was designed and implemented with Java in Intellij IDEA, where all Matlab codes were compiled into Java Libraries. As the result, this system is a Java Package that can be executed on any computer which has the Java Virtual Machine and a Matlab compiler installed on it.

For the evaluation, we test the proposed DVT method and the developed system on a sitcom episode and a TV show on which the face-TLD method was also validated [21]. In addition, we also conduct experiments on one short type of TV show to visualize the results in a qualitative trend.

We compare the proposed DVT method with the standard TLD and face-TLD by testing them on the sitcom IT-Crowd (first series, first episode). The episode duration is 1418 seconds, with the frame rate of 29 frame per second. Table 1 provides the performance comparison of the three methods in term of precision and recall measures as described in [2]. The developed system with DVT method is able to detect the query face through the whole video. The overall recall is 75%, and the precision is 92%, both of which are much larger than TLD and face-TLD methods. Moreover, for the initialization, the developed system is initialized with a bounding box on the character Roy at any desired time within the video, while the TLD and face-TLD methods need to perform the initialization on the first appearance of the character.

A visualization of the user interface of the DVT system is also given in Figure 4. In the provided graphical user interface, the user selects a query face in some desired frame by drawing a bounding box around the target face. Thereafter, the tracking system tracks the query face within the entire video regardless of the fact that the selected query face might not be the first appearance of the target in the video. After the tracking is completed, the GUI shows a time-bar which represents the duration of the video with the highlighted parts in cyan which indicate the time slots when the query person appears in the video. The query face is also bounded in a yellow box in each frame the query person appears for better visualization.

In all experiments, the similarity threshold (Algorithm 1, line 2) is set to 70%. The number of skipping frames (Algorithm 1, line 4) can be specified by the user through the GUI, where the default setting is 60. The number of skipping frames indicates how many frames are skipped by the system when face verification is failed. Although decreasing this number will lead to an increase in recall value, it will increase the running time of the system which is in contradiction to the goal of tracking in real-time. A sample of DVT output sequence is also given in Fig. 3.

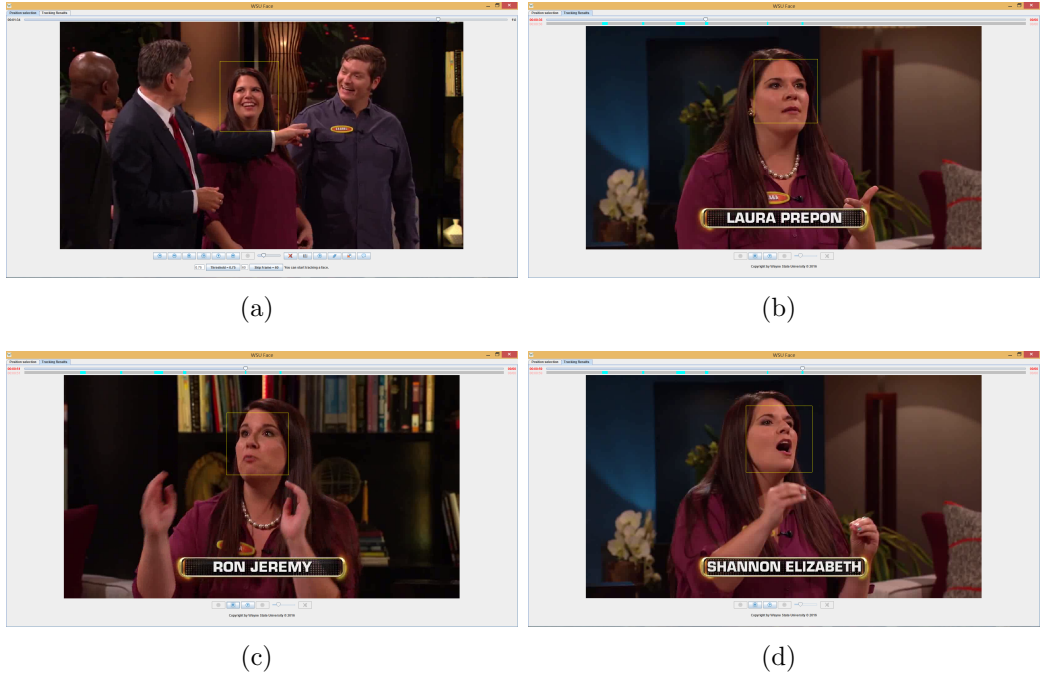

(a) (b) (c) (d)

Figure 2: A visualization of the user interface of the developed long-term face tracking system, (a) the user selects a query face in some desired frame, (b, c, d) the tracking system tracks the query face within the entire video. The time-bar represents the duration of the video, and the highlighted parts in cyan indicate the time slots when the query person appears in the video. The query face is also bounded in a yellow box in each frame the query person appears.

Table 1: The comparison between TLD, Face-TLD and the proposed DVT method in terms of precision and recall.

| Method | Character Roy | |
|---|---|---|
| | Precision | Recall |
| TLD | 0.70 | 0.37 |
| Face-TLD | 0.75 | 0.54 |
| DVT (the proposed) | 0.95 | 0.75 |

## 5. summary

This paper presents a deep-learning based detection-verification-tracking method and develops a system for long-term tracking of human faces in unconstrained videos.

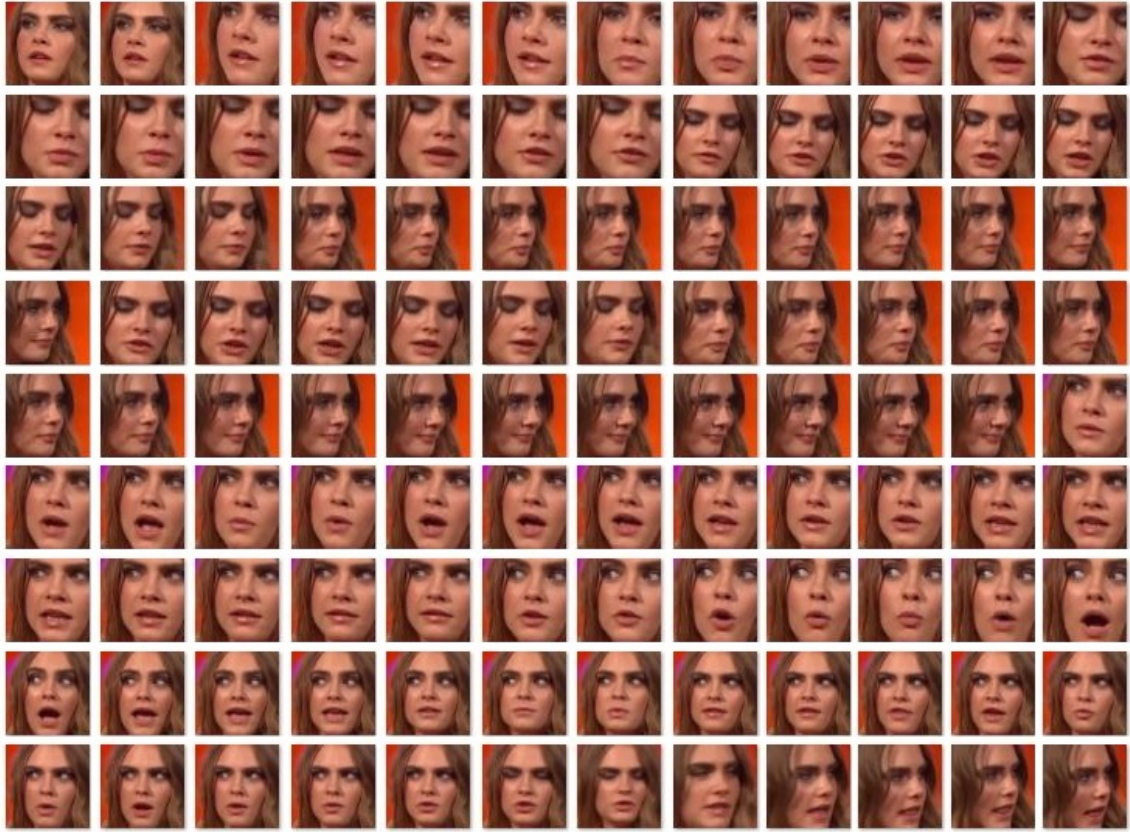

Figure 3: A sample of DVT output sequence.

The system employs face detection and face verification to boost the performance of long-term tracking. By testing the system with DVT method on a sitcom episode, a TV show, and other types of videos, its efficacy is validated, and the system is promising to be used in real-time applications.

# References

[1] Z. Kalal, K. Mikolajczyk, J. Matas, Tracking-learning-detection, Pattern Analysis and Machine Intelligence, IEEE Transactions on 34 (7) (2012) 1409–1422.

[2] Z. Kalal, K. Mikolajczyk, J. Matas, Face-tld: Tracking-learning-detection applied to faces, in: Image Processing (ICIP), 2010 17th IEEE International Conference on, IEEE, 2010, pp. 3789–3792.

[3] J. Supancic, D. Ramanan, Self-paced learning for long-term tracking, in: Proceedings of the IEEE Conference on Computer Vision and Pattern Recognition, 2013, pp. 2379–2386.

[4] Y. Li, J. Zhu, S. C. Hoi, Reliable patch trackers robust visual tracking by exploiting reliable patches, in: Proceedings of the IEEE Conference on Computer Vision and Pattern Recognition, 2015, pp. 353–361.

[5] A. W. Smeulders, D. M. Chu, R. Cucchiara, S. Calderara, A. Dehghan, M. Shah, Visual tracking: An experimental survey, Pattern Analysis and Machine Intelligence, IEEE Transactions on 36 (7) (2014) 1442–1468.

[6] R. Wang, H. Dong, T. X. Han, L. Mei, Robust tracking via monocular active vision for an intelligent teaching system, The Visual Computer (2016) 1–16.

[7] O. Williams, A. Blake, R. Cipolla, Sparse bayesian learning for efficient visual tracking, Pattern Analysis and Machine Intelligence, IEEE Transactions on 27 (8) (2005) 1292–1304.

[8] Y. Li, H. Ai, T. Yamashita, S. Lao, M. Kawade, Tracking in low frame rate video: A cascade particle filter with discriminative observers of different life spans, Pattern Analysis and Machine Intelligence, IEEE Transactions on 30 (10) (2008) 1728–1740.

[9] H. Grabner, C. Leistner, H. Bischof, Semi-supervised on-line boosting for robust tracking, in: Computer Vision–ECCV 2008, Springer, 2008, pp. 234–247.

[10] B. Babenko, M.-H. Yang, S. Belongie, Visual tracking with online multiple instance learning, in: Computer Vision and Pattern Recognition, 2009. CVPR 2009. IEEE Conference on, IEEE, 2009, pp. 983–990.

[11] Z. Kalal, J. Matas, K. Mikolajczyk, Weighted sampling for large-scale boosting., in: BMVC, 2008, pp. 1–10.

[12] W.-L. Zheng, S.-C. Shen, B.-L. Lu, Online depth image-based object tracking with sparse representation and object detection, Neural Processing Letters (2016) 1–14.

[13] S. Zafeiriou, C. Zhang, Z. Zhang, A survey on face detection in the wild: past, present and future, Computer Vision and Image Understanding 138 (2015) 1–24.

[14] S. Yang, P. Luo, C.-C. Loy, X. Tang, From facial parts responses to face detection: A deep learning approach, in: Proceedings of the IEEE International Conference on Computer Vision, 2015, pp. 3676–3684.

[15] H. Li, Z. Lin, X. Shen, J. Brandt, G. Hua, A convolutional neural network cascade for face detection, in: Proceedings of the IEEE Conference on Computer Vision and Pattern Recognition, 2015, pp. 5325–5334.

[16] O. M. Parkhi, A. Vedaldi, A. Zisserman, Deep face recognition, Proceedings of the British Machine Vision 1 (3) (2015) 6.

[17] Y. Sun, Y. Chen, X. Wang, X. Tang, Deep learning face representation by joint identification-verification, in: Advances in Neural Information Processing Systems, 2014, pp. 1988–1996.

[18] F. Schroff, D. Kalenichenko, J. Philbin, Facenet: A unified embedding for face recognition and clustering, in: Proceedings of the IEEE Conference on Computer Vision and Pattern Recognition, 2015, pp. 815–823.

[19] J. Kwon, K. M. Lee, Highly nonrigid object tracking via patch-based dynamic appearance modeling, Pattern Analysis and Machine Intelligence, IEEE Transactions on 35 (10) (2013) 2427–2441.

[20] M. Godec, P. M. Roth, H. Bischof, Hough-based tracking of non-rigid objects, Computer Vision and Image Understanding 117 (10) (2013) 1245–1256.

[21] S. Dubuisson, C. Gonzales, A survey of datasets for visual tracking, Machine Vision and Applications 27 (1) (2016) 23–52.